\documentclass[]{article}  % Comment this line out
\usepackage{neurips_rl2019}

\usepackage{graphicx}                       % Necessary to include graphics
\usepackage{graphics}                       % To include pdf, bitmapped graphics files
\usepackage{epsfig}                         % To include eps files
\usepackage[tight,footnotesize]{subfigure}  % Create subfigures, ie 1A, 1B

\usepackage{bm}

\usepackage{cancel}
\usepackage{xifthen}
\newcommand\Ccancel[2][black]{\renewcommand\CancelColor{\color{#1}}\cancel{#2}}
\newcommand{\mz}[2][]{%
\ifthenelse{\isempty{#1}{}}{}{\Ccancel[cyan]{#1}}%
{\color{cyan} #2}%
}

\newcommand{\jim}[2][]{%
\ifthenelse{\isempty{#1}{}}{}{\Ccancel[olive]{#1}}%
{\color{olive} #2}%
}

\usepackage{amssymb,amsmath}    %\usepackage{amslatex}%
\usepackage{mdwmath}
\usepackage{commath}   % Used for \norm \abs
\usepackage{eqparbox}
\usepackage{mathtools}
\usepackage[utf8]{inputenc} % Used for theorem/corollary/lemma
\usepackage[english]{babel}
\usepackage{amsmath,amsfonts,amssymb}

\newcommand{\beq}{\begin{equation}}
\newcommand{\eeq}{\end{equation}}
\newcommand{\bear}{\begin{eqnarray}}
\newcommand{\bears}{\begin{eqnarray*}}
\newcommand{\eear}{\end{eqnarray}}
\newcommand{\eears}{\end{eqnarray*}}
\newcommand{\bdm}{\begin{displaymath}}
\newcommand{\edm}{\end{displaymath}}
\newcommand{\lba}{\left[\begin{array}}
\newcommand{\ear}{\end{array}\right]}

\usepackage{stfloats}                       % Create Navigation Titles in PDF
\usepackage{url} % bibtex
\usepackage{hyperref}
\usepackage[T1]{fontenc} % Underscores (to better visualize them using the appropriate font)
\usepackage{tabularx}
\usepackage{multirow}
\usepackage[table]{xcolor}
\usepackage{longtable}
\usepackage{booktabs} 			% professional-quality tables and \midrule
\usepackage{xcolor,colortbl}    % Color columns, use with: \usepackage{array}

\usepackage{authblk}
\title{Towards More Sample Efficiency in\\ Reinforcement Learning with Data Augmentation}
\author[1]{Yijiong Lin}
\author[1]{Jiancong Huang}
\author[2]{Matthieu Zimmer}
\author[1]{Yisheng Guan}
\author[1]{Juan Rojas}
\author[2]{Paul Weng}
\affil[1]{Guangdong University of Technology}
\affil[2]{Shanghai Jiao Tong University}
\date{}                     %% if you don't need date to appear
\begin{document}
\maketitle
%----------------------------------------------------------------------------------
%----------------------------------------------------------------------------------
\begin{abstract}
%----------------------------------------------------------------------------------
Deep reinforcement learning (DRL) is a promising approach for adaptive robot control, but its current application to robotics is currently hindered by high sample requirements.
We propose two novel data augmentation techniques for DRL in order to reuse more efficiently observed data.
The first one called Kaleidoscope Experience Replay exploits reflectional symmetries, while the second called Goal-augmented Experience Replay takes advantage of lax goal definitions.
Our preliminary experimental results show a large increase in learning speed.
%----------------------------------------------------------------------------------
%----------------------------------------------------------------------------------
\end{abstract}
%----------------------------------------------------------------------------------
\section{Introduction}\label{sec:Intro}
%----------------------------------------------------------------------------------
% Background
Deep reinforcement learning (DRL) has demonstrated great promise in recent years \citep{mnih2015human, alphago}.
However, despite being shown to be a viable approach in robotics \citep{pmlr-v87-kalashnikov18a, openai2018learning}, DRL still suffers from high sample complexity in practice---an acute issue in robot learning.

% Related Works
Given how critical this issue is, many diverse propositions have been presented. For brevity, we only recall the most  related to our work.
% Better Observation Reuse
A first idea is to better utilize observed samples, e.g., memory replay \citep{lin1992self} or hindsight experience replay (HER) \citep{Andrychowicz2017HindsightReplay}.
Although better observation reuse does not reduce sample requirements in a DRL algorithm, it decreases the number of actual interactions with the environment, which is the most important factor in robot learning.
% Prior Domain Knowledge
Another idea is to exploit any a priori domain knowledge one may have (e.g., symmetry \citep{kidzinski2018learning}) to support learning. Besides, a robot is generally expected to solve not only one fixed task, but multiple related ones.
% Multi-task RL
Multi-task reinforcement learning \citep{Plappert2018Multi-GoalResearch} is considered beneficial as it would not be feasible to repeatedly solve each encountered task tabula rasa.
% Transfer Learning
Finally, in order to avoid or reduce learning on an actual robot, recent works have investigated how policies learned in a simulator can be transferred to a real robot \citep{tobin2017domain}.

% Research Question/Contribution
In this work, in order to further reduce the requirement on the number of actual samples, we propose two novel data augmentation methods that better reuse the samples observed in the true environment by exploiting the symmetries (i.e., any invariant transformation) of the problem. For simplicity, we present them in the same set-up as HER \citep{Andrychowicz2017HindsightReplay}, although the techniques can be instantiated in other settings with any DRL algorithm based on memory replay. 
%Our two proposed techniques can, interestingly, also be combined in a synergistic manner with HER as we demonstrate in our preliminary experiments.
%However, the first technique is more general could be employed in a more general context than HER.

% Methods
The first technique, Kaleidoscope Experience Replay (KER), is based on reflectional symmetry, it posits that trajectories in robotic spaces can enjoy multiple reflections and remain in the valid workspace. 
%Namely, trajectories can be repeatedly transformed such that valid trajectories are mapped to valid trajectories. 
Namely, for a given robotic problem, there is a set of reflections %(whose validity is problem dependent e.g., robot, task, domain) 
that can generate from any valid observed trajectory many new artificial valid ones for training.
For concreteness, in this paper we focus on reflections with respect to hyperplanes. 

The second technique, Goal-augmented Experience Replay (GER), can be seen as a generalization of HER:
any artificial goal $g$ generated by HER can be instead replaced by a random goal sampled in a small ball around $g$.
This idea takes advantage of 
tasks where success is defined as reaching a final pose within a distance of the goal set by a threshold (such tasks are common in robotics). 
%The second technique, Goal-augmented Experience Replay (GER), applies to tasks where success is defined as reaching a final pose within a distance of the goal set by a threshold (such tasks are common in robotics). 
%Here, the idea is to augment observed successful trajectories with new trajectories whose goals are replaced by a new random position near the end position of the original trajectory.
Here, successful trajectories are augmented using the invariant transformation that consists in changing actual goals to random close goals. %whose distance to the final pose is within the fixed threshold.

%% TODO: I would discuss results/impact here to draw the reader's attention to the significance of our work. If space is an issue, I would make sure to include this section over the "organization" paragraph. 

% Organization
In Sec. \ref{sec:back}, we present this work's setup.  Sec. \ref{sec:related} introduces related work. Sec. \ref{sec:aer} details our two data augmentation techniques. Sec. \ref{sec:expe} presents preliminary results; and Sec. \ref{sec:conclusion} highlights key lessons.
%----------------------------------------------------------------------------------
\section{Background} \label{sec:back}
%----------------------------------------------------------------------------------
% Multiple goals
In this work, we consider robotic tasks that are modeled as multi-goal Markov decision processes \citep{schaul2015universal} with continuous state and action spaces: $\langle \mathcal S, \mathcal A, \mathcal G, T, R, p, \gamma \rangle$ where $\mathcal S$ is a continuous state space, $\mathcal A$ is a continuous action space, $\mathcal G$ is a set of goals, $T$ is the unknown transition function that describes the effects of actions, $R(s, a, s', g)$ is the immediate reward when reaching state $s' \in \mathcal S$ after performing action $a \in \mathcal A$ in state $s \in \mathcal S$ if the goal were $g \in \mathcal G$.
Finally, $p$ is a joint probability distribution over initial states and initial goals, and $\gamma \in (0, 1)$ is a discount factor.
In this framework, the robot learning problem corresponds to an RL problem that aims at obtaining a policy $\pi : \mathcal S \times \mathcal G \to \mathcal A$ such that the expected discounted sum of rewards is maximized for any given goal.

% Sparse Rewards
When the reward function is sparse, as assumed here, this RL problem is particularly hard to solve.
In particular, we consider here reward functions that are described as follows:
$R(s, a, s', g) = \bm 1[ d(s', g) \le \epsilon_R] -1$ where $\bm 1$ is the indicator function, $d$ is a distance, and $\epsilon_R>0$ is a fixed threshold.

% HER
To tackle this issue, \cite{Andrychowicz2017HindsightReplay} proposed HER, which is based on the following principle:
Any trajectory that failed to reach its goal still carries useful information; it has at least reached the states of its trajectory path. Using this natural and powerful idea, memory replay can be augmented with the failed trajectories by changing their goals in \textit{hindsight}.

%Statistics
%----------------------------------------------------------------------------------
\section{Related Work} \label{sec:related}
%----------------------------------------------------------------------------------
HER \citep{Andrychowicz2017HindsightReplay,Plappert2018Multi-GoalResearch} has been extended in various ways.
% PER
Prioritized replay was incorporated in HER to learn from more valuable episodes with higher priority \citep{Zhao2018Energy-BasedPrioritization}.
% Dynamic Goals
In \citep{Fang2019DHER:Replay}, HER was generalized to deal with dynamic goals. In \citep{Gerken2019ContinuousControllers}, a variant of HER was also investigated where completely random goals replace achieved goals and in \citep{Rauber2019HindsightGradients}, it was adapted to work with on-policy RL algorithms. All these extensions are orthogonal to our work and could easily be combined with KER. We leave these for future work.

% Symmetry
Symmetry has been considered in MDPs \citep{Zinkevich2001SymmetryLearning} and RL \citep{Kamal2008ReinforcementStates,Agostini2009ExploitingSpaces,Mahajan2017SymmetryLearning,Kidzinski2018LearningEnvironments,Amadio2019ExploitingTasks}. It can be known a priori or learned \citep{Mahajan2017SymmetryLearning}. In this work, we assume the former, which is reasonable in many robotics tasks. 
% Aggregate Symmetry States
A natural approach to exploit symmetry in sequential decision-making is by aggregating states that satisfy an equivalence relation induced by a symmetry \citep{Zinkevich2001SymmetryLearning,Kamal2008ReinforcementStates}.
% Symmetry in Policy Representation
Another related approach takes into account symmetry in the policy representation \citep{Amadio2019ExploitingTasks}. Doing so reduces representation size and generally leads to faster solution times. However, the state-aggregated representation may be difficult to recover, especially if many symmetries are considered simultaneously.
% Symmetry during Training
Still another approach is to use symmetry during training instead.
One simple idea is to learn the Q-function by performing an additional symmetrical update \citep{Agostini2009ExploitingSpaces}.
% Train with reflections. 
Another method is to augment the training points with their reflections \citep{Kidzinski2018LearningEnvironments}.
In this paper, we generalize further this idea as a data augmentation technique where many symmetries can be considered and pairs of symmetrical updates do not need to be simultaneously applied.

While, to the best of our knowledge, data augmentation has not been considered much to accelerate learning in RL, it has been used extensively and with great success in machine learning \citep{Baird1992DocumentModels} and more so in deep learning \citep{KrizhevskyImagenetNetworks}.
Interestingly, symmetries can also be exploited in neural network architecture design \citep{Gens2014DeepNetworks}. %, reflecting the two previous approaches in MDP and RL.
However, in our case, the integration of symmetry in deep networks will be left as future work.
%--------------------------------------------------------    
\section{Data Augmentation for RL}\label{sec:aer}
%--------------------------------------------------------    
To reduce the number of interactions with the real environment our goal is to generate artificial training data based on actual trajectories collected during the robot's learning.

\begin{figure}[t]
  \centering
    \includegraphics[width=.85\linewidth]{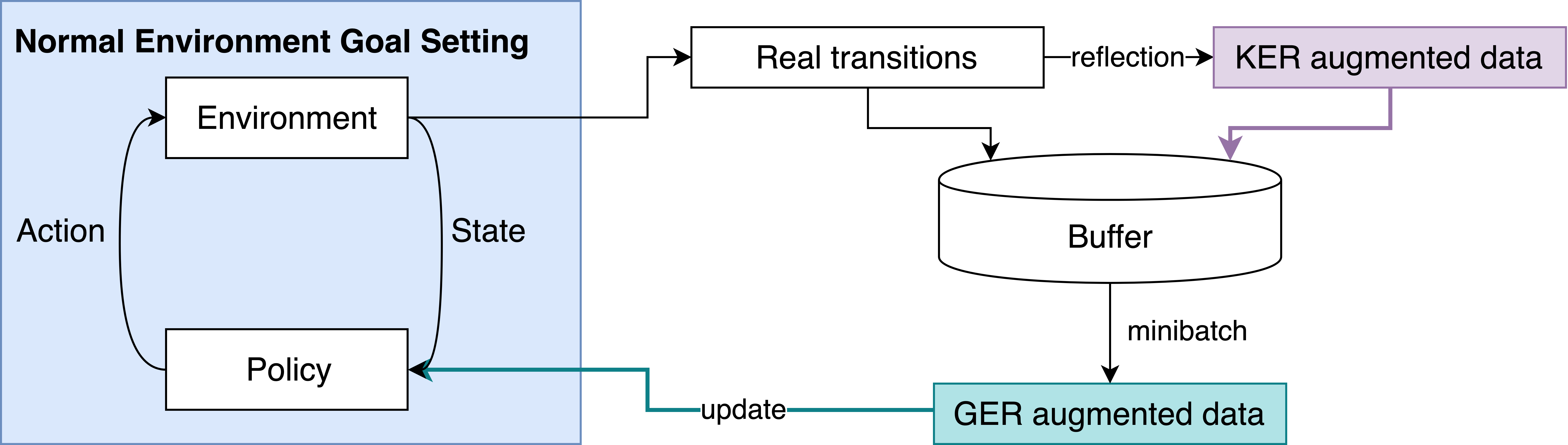}
    \caption{Framework overview: real and symmetrically transformed transitions are stored in the replay buffer.  %80\% of the experiences are augmented by HER, and these along with the other 20\% are further augmented by GER before updating the policy.
    Sampled minibatches are then augmented with GER before updating the policy.
    }
  \label{fig:architecture}
\end{figure}

Our architecture leverages our two proposed techniques, Kaleidoscope experience replay (KER) and Goal-Augmented Experience Replay (GER). 
While the two methods are combined here, only one of them could be used instead. An overview of our architecture is illustrated in Fig. \ref{fig:architecture}.
%--------------------------------------------------------    
%-------------------------------------------------------- 
\subsection{Kaleidoscope Experience Replay (KER)}\label{subsec:ker}
\begin{figure}[!h]
  \centering
  \hfill\begin{minipage}{0.45\textwidth}
    \includegraphics[width=.8\linewidth]{pics/symmetry_plane_whole2.png}
   \end{minipage}
   \begin{minipage}{0.45\textwidth}
    \includegraphics[width=1.\linewidth]{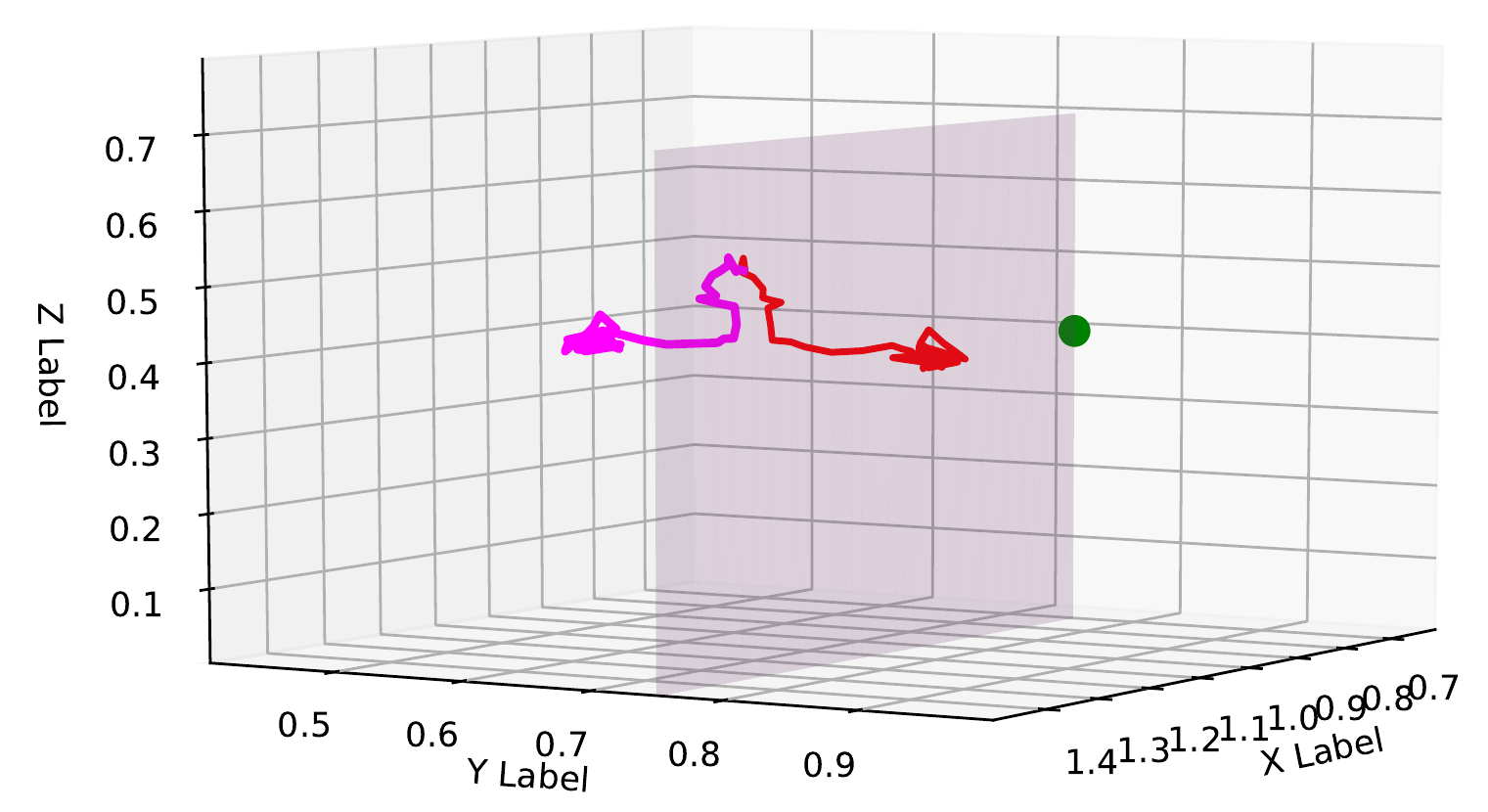}\hfill
    \end{minipage}\hspace{0.7cm}
    \caption{Kaleidoscope Experience Replay leverages natural symmetry. Valid trajectories are reflected with respect to plane $xoz$, where the latter can itself be rotated by some $\theta_z$ along axis $\vec{z}$.}
  \label{fig:ker}
\end{figure}

%-------------------------------------------------------- 
KER uses reflectional symmetry\footnote{Though more general invariant transformations could also be used in place of reflectional symmetry.}. Consider a 3D workspace with a bisecting plane $xoz$ as shown in Fig. \ref{fig:ker}. If a valid trajectory is generated in the workspace (blue in Fig. \ref{fig:ker}), natural symmetry would then yield a new valid trajectory reflected on this plane. More generally, the $xoz$ plane may be rotated by some angle $\theta_z$ along axis $\vec{z}$ and still define an invariant symmetry for the robotic task.
%--------------------------------------------------------    
%-------------------------------------------------------- 
We can now precisely define KER, which amounts to augmenting any original trajectory with a certain number of random symmetries. A random symmetry is a reflectional symmetry with respect to the $xoz$ plane after it has been rotated by a random angle about the $\vec{z}$-axis.

Note that instead of storing the reflected trajectories in the replay buffer, the random symmetries could be instead applied to sampled minibatches. This approach was tried previously for single-symmetry scenarios \citep{kidzinski2018learning}. Doing so, however, is more computationally taxing as transitions are reflected every time they are sampled and more significantly leads to lower performance\footnote{Please visit our project page for this and more supplementary information (not referenced until after review).}. Our conjecture is that such approach leads to a lower diversity in the minibatches.
%-------------------------------------------------------- 
\subsection{Goal-Augmented Experience Replay (GER)}
%-------------------------------------------------------- 
GER exploits the formulation of any reward function that defines a successful trajectory as one whose end position is within a small radial threshold (a ball) centered around the goal. When the robot obtains a successful trajectory, we therefore know that it can in fact be considered successful for any goal within a ball centered around its end position.
Based on this observation, GER augments successful trajectories by replacing the original goal with a random goal sampled within that ball.
This ball can be formally described as $\{ g \in \mathcal G | d(s_f, g) \le \epsilon  \}$ where $s_f$ is the final state reached in the original trajectory and $\epsilon < \epsilon_R$ is a threshold, which does not have to be constant for each application of GER.
Therefore, GER can be seen as a generalization of HER and can be implemented in the same fashion.
This is why in our architecture, GER is applied on minibatches, like HER.

%----------------------------------------------------------------------------------
\section{Preliminary Experimental Results} \label{sec:expe}
%----------------------------------------------------------------------------------

Our experimental evaluation is performed according to the HER formulation \citep{Andrychowicz2017HindsightReplay}. Namely, we use a simulated 7-DOF Fetch Robotics arm trained with DDPG on the pushing, sliding, and pick-and-place tasks.
 
\begin{figure}[!h]
  \centering
    \includegraphics[width=1.\linewidth]{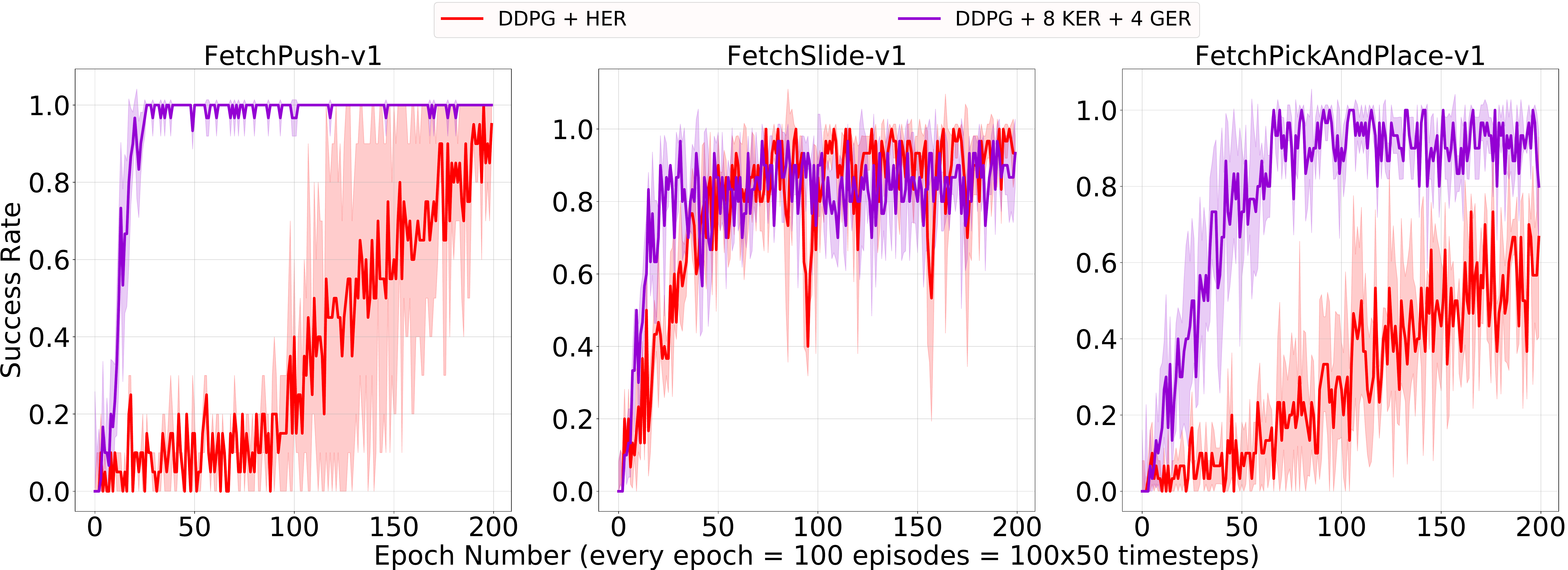}
%   \begin{minipage}{0.325\textwidth}
%     \includegraphics[width=1.0\linewidth]{pics/FetchPickAndPlace-v1.pdf}
%     \end{minipage}\hfill
%   \begin{minipage}{0.325\textwidth}
%     \includegraphics[width=1.0\linewidth]{pics/FetchSlide-v1.pdf}
%     \end{minipage}%
    \caption{Comparison of vanilla HER with the combination of 8 KER symmetries and 4 GERs.}
  \label{fig:results}
\end{figure}

We design our experiments to demonstrate the effectiveness of our propositions and final combination, which uses $n=8$ random symmetries for KER and $4$ applications of GER (where one of them uses a threshold equal to zero in order to also take full advantage of realized goals).
We now present some initial experimental results.
As shown in Fig. \ref{fig:results}, our method vastly improves the learning speed compared to vanilla HER.

In our experiments, we have observed that performance is monotonic with respect to the number $n$ of random symmetries, although the gain diminishes for larger $n$.
Similar observations can be made for the number of applications of GER.

%----------------------------------------------------------------------------------
% %----------------------------------------------------------------------------------
% \subsection{KER-based Experiments}
% %----------------------------------------------------------------------------------
% Here we show that adding KER to HER drastically accelerate learning.

% Add some figures and explain.
% %----------------------------------------------------------------------------------
% \subsection{GER-based Experiments}
% %----------------------------------------------------------------------------------
% Here we show that augmenting with random goals also improves the performance.

% Add some figures and explain.

% \subsection{How many symmetries should we use in KER?}

% Here we present the plot "performance vs number of symmetries".

% Talk about simulated Baxter? and transfer to real Baxter?

%----------------------------------------------------------------------------------
\section{Conclusion} \label{sec:conclusion}
%----------------------------------------------------------------------------------
We proposed two novel data augmentation techniques KER and GER to amplify the efficiency of observed samples in a memory replay mechanism. KER exploited reflectional symmetry in the valid workspace (though in general it could be employed with other types of symmetries). 
% Talk about KERs impact/significance/key lessons learned.
GER, as an extension of HER, is specific to goal-oriented tasks where success is defined in terms of a thresholded distance.
The combination of these techniques greatly accelerated learning as demonstrated in our experiments.

Our next step is to use our method to solve the same tasks with a simulated Baxter robot, and then transfer to a real Baxter using a sim2real methodology.
Furthermore, we aim at extending our proposition to other types of symmetries and to other robotic tasks as well.
%More investigation is needed to better understand how to exploit those techniques most adequately; in particular, to avoid the performance drop observed when too many artificial trajectories were added.
%In future work, we will also consider more diverse symmetries.
\clearpage
%----------------------------------------------------------------------------------
%% BIBLIOGRAPHY
%----------------------------------------------------------------------------------
\bibliographystyle{named}
\bibliography{IEEEabrv,references}

\begin{thebibliography}{}

\bibitem[\protect\citeauthoryear{Agostini and
  Celaya}{2009}]{Agostini2009ExploitingSpaces}
Alejandro Agostini and Enric Celaya.
\newblock {Exploiting Domain Symmetries in Reinforcement Learning with
  Continuous State and Action Spaces}.
\newblock In {\em ICMLA}, 2009.

\bibitem[\protect\citeauthoryear{Amadio \bgroup \em et al.\egroup
  }{2019}]{Amadio2019ExploitingTasks}
Fabio Amadio, Adria Colome, and Carme Torras.
\newblock {Exploiting Symmetries in Reinforcement Learning of Bimanual Robotic
  Tasks}.
\newblock {\em IEEE Robotics and Automation Letters}, 4(2):1838--1845, 4 2019.

\bibitem[\protect\citeauthoryear{Andrychowicz \bgroup \em et al.\egroup
  }{2017}]{Andrychowicz2017HindsightReplay}
Marcin Andrychowicz, Filip Wolski, Alex Ray, Jonas Schneider, Rachel Fong,
  Peter Welinder, Bob McGrew, Josh Tobin, Pieter Abbeel, and Wojciech Zaremba.
\newblock {Hindsight experience replay}.
\newblock In {\em Advances in Neural Information Processing Systems}, pages
  5049--5059, 2017.

\bibitem[\protect\citeauthoryear{Baird}{1992}]{Baird1992DocumentModels}
Henry~S. Baird.
\newblock {Document Image Defect Models}.
\newblock In {\em Structured Document Image Analysis}, pages 546--556.
  Springer, 1992.

\bibitem[\protect\citeauthoryear{Fang \bgroup \em et al.\egroup
  }{2019}]{Fang2019DHER:Replay}
Meng Fang, Cheng Zhou, Bei Shi, Boqing Gong, Jia Xu, and Tong Zhang.
\newblock {DHER: Hindsight Experience Replay}.
\newblock In {\em ICLR}, 2019.

\bibitem[\protect\citeauthoryear{Gens and
  Domingos}{2014}]{Gens2014DeepNetworks}
Robert Gens and Pedro Domingos.
\newblock {Deep symmetry networks}.
\newblock In {\em Advances in Neural Information Processing Systems}, pages
  2537--2545, 2014.

\bibitem[\protect\citeauthoryear{Gerken and
  Spranger}{2019}]{Gerken2019ContinuousControllers}
Andreas Gerken and Michael Spranger.
\newblock {Continuous Value Iteration (CVI) Reinforcement Learning and
  Imaginary Experience Replay (IER) For Learning Multi-Goal, Continuous Action
  and State Space Controllers}.
\newblock In {\em 2019 International Conference on Robotics and Automation
  (ICRA)}, pages 7173--7179. IEEE, 5 2019.

\bibitem[\protect\citeauthoryear{Kalashnikov \bgroup \em et al.\egroup
  }{2018}]{pmlr-v87-kalashnikov18a}
Dmitry Kalashnikov, Alex Irpan, Peter Pastor, Julian Ibarz, Alexander Herzog,
  Eric Jang, Deirdre Quillen, Ethan Holly, Mrinal Kalakrishnan, Vincent
  Vanhoucke, and Sergey Levine.
\newblock Scalable deep reinforcement learning for vision-based robotic
  manipulation.
\newblock In Aude Billard, Anca Dragan, Jan Peters, and Jun Morimoto, editors,
  {\em Proceedings of The 2nd Conference on Robot Learning}, volume~87 of {\em
  Proceedings of Machine Learning Research}, pages 651--673. PMLR, 29--31 Oct
  2018.

\bibitem[\protect\citeauthoryear{Kamal and
  Murata}{2008}]{Kamal2008ReinforcementStates}
M.A.S. Kamal and Junichi Murata.
\newblock {Reinforcement learning for problems with symmetrical restricted
  states}.
\newblock {\em Robotics and Autonomous Systems}, 56(9):717--727, 9 2008.

\bibitem[\protect\citeauthoryear{Kidzi{\'n}ski \bgroup \em et al.\egroup
  }{2018a}]{kidzinski2018learning}
{\L}ukasz Kidzi{\'n}ski, Sharada~Prasanna Mohanty, Carmichael~F Ong, Zhewei
  Huang, Shuchang Zhou, Anton Pechenko, Adam Stelmaszczyk, Piotr Jarosik,
  Mikhail Pavlov, Sergey Kolesnikov, et~al.
\newblock Learning to run challenge solutions: Adapting reinforcement learning
  methods for neuromusculoskeletal environments.
\newblock In {\em The NIPS'17 Competition: Building Intelligent Systems}, pages
  121--153. Springer, 2018.

\bibitem[\protect\citeauthoryear{Kidzi{\'{n}}ski \bgroup \em et al.\egroup
  }{2018b}]{Kidzinski2018LearningEnvironments}
Łukasz Kidzi{\'{n}}ski, Sharada~Prasanna Mohanty, Carmichael~F. Ong, Zhewei
  Huang, Shuchang Zhou, Anton Pechenko, Adam Stelmaszczyk, Piotr Jarosik,
  Mikhail Pavlov, Sergey Kolesnikov, Sergey Plis, Zhibo Chen, Zhizheng Zhang,
  Jiale Chen, Jun Shi, Zhuobin Zheng, Chun Yuan, Zhihui Lin, Henryk
  Michalewski, Piotr Milos, Blazej Osinski, Andrew Melnik, Malte Schilling,
  Helge Ritter, Sean~F. Carroll, Jennifer Hicks, Sergey Levine, Marcel
  Salath{\'{e}}, and Scott Delp.
\newblock {Learning to Run Challenge Solutions: Adapting Reinforcement Learning
  Methods for Neuromusculoskeletal Environments}.
\newblock In {\em The NIPS '17 Competition: Building Intelligent Systems},
  pages 121--153. Springer, 2018.

\bibitem[\protect\citeauthoryear{Krizhevsky \bgroup \em et al.\egroup
  }{2012}]{KrizhevskyImagenetNetworks}
Alex Krizhevsky, Ilya Sutskever, and Geoffrey~E Hinton.
\newblock Imagenet classification with deep convolutional neural networks.
\newblock In {\em Advances in neural information processing systems}, pages
  1097--1105, 2012.

\bibitem[\protect\citeauthoryear{Lin}{1992}]{lin1992self}
Long-Ji Lin.
\newblock Self-improving reactive agents based on reinforcement learning,
  planning and teaching.
\newblock {\em Machine learning}, 8(3-4):293--321, 1992.

\bibitem[\protect\citeauthoryear{Mahajan and
  Tulabandhula}{2017}]{Mahajan2017SymmetryLearning}
Anuj Mahajan and Theja Tulabandhula.
\newblock Symmetry learning for function approximation in reinforcement
  learning.
\newblock {\em arXiv preprint arXiv:1706.02999}, 2017.

\bibitem[\protect\citeauthoryear{Mnih \bgroup \em et al.\egroup
  }{2015}]{mnih2015human}
Volodymyr Mnih, Koray Kavukcuoglu, David Silver, Andrei~A Rusu, Joel Veness,
  Marc~G Bellemare, Alex Graves, Martin Riedmiller, Andreas~K Fidjeland, Georg
  Ostrovski, et~al.
\newblock Human-level control through deep reinforcement learning.
\newblock {\em Nature}, 518(7540):529, 2015.

\bibitem[\protect\citeauthoryear{OpenAI \bgroup \em et al.\egroup
  }{2018}]{openai2018learning}
OpenAI, Marcin Andrychowicz, Bowen Baker, Maciek Chociej, Rafał Józefowicz,
  Bob McGrew, Jakub Pachocki, Arthur Petron, Matthias Plappert, Glenn Powell,
  Alex Ray, Jonas Schneider, Szymon Sidor, Josh Tobin, Peter Welinder, Lilian
  Weng, and Wojciech Zaremba.
\newblock Learning dexterous in-hand manipulation.
\newblock {\em CoRR}, 2018.

\bibitem[\protect\citeauthoryear{Plappert \bgroup \em et al.\egroup
  }{2018}]{Plappert2018Multi-GoalResearch}
Matthias Plappert, Marcin Andrychowicz, Alex Ray, Bob McGrew, Bowen Baker,
  Glenn Powell, Jonas Schneider, Josh Tobin, Maciek Chociej, Peter Welinder,
  et~al.
\newblock Multi-goal reinforcement learning: Challenging robotics environments
  and request for research.
\newblock {\em arXiv preprint arXiv:1802.09464}, 2018.

\bibitem[\protect\citeauthoryear{Rauber \bgroup \em et al.\egroup
  }{2017}]{Rauber2019HindsightGradients}
Paulo Rauber, Avinash Ummadisingu, Filipe Mutz, and Juergen Schmidhuber.
\newblock Hindsight policy gradients.
\newblock {\em arXiv preprint arXiv:1711.06006}, 2017.

\bibitem[\protect\citeauthoryear{Schaul \bgroup \em et al.\egroup
  }{2015}]{schaul2015universal}
Tom Schaul, Daniel Horgan, Karol Gregor, and David Silver.
\newblock Universal value function approximators.
\newblock In {\em International Conference on Machine Learning}, pages
  1312--1320, 2015.

\bibitem[\protect\citeauthoryear{Silver \bgroup \em et al.\egroup
  }{2016}]{alphago}
David Silver, Aja Huang, Chris~J Maddison, Arthur Guez, Laurent Sifre, George
  Van Den~Driessche, Julian Schrittwieser, Ioannis Antonoglou, Veda
  Panneershelvam, Marc Lanctot, et~al.
\newblock Mastering the game of go with deep neural networks and tree search.
\newblock {\em nature}, 529(7587):484, 2016.

\bibitem[\protect\citeauthoryear{Tobin \bgroup \em et al.\egroup
  }{2017}]{tobin2017domain}
Josh Tobin, Rachel Fong, Alex Ray, Jonas Schneider, Wojciech Zaremba, and
  Pieter Abbeel.
\newblock Domain randomization for transferring deep neural networks from
  simulation to the real world.
\newblock In {\em 2017 IEEE/RSJ International Conference on Intelligent Robots
  and Systems (IROS)}, pages 23--30. IEEE, 2017.

\bibitem[\protect\citeauthoryear{Zhao and
  Tresp}{2018}]{Zhao2018Energy-BasedPrioritization}
Rui Zhao and Volker Tresp.
\newblock {Energy-Based Hindsight Experience Prioritization}.
\newblock In {\em coRL}, 2018.

\bibitem[\protect\citeauthoryear{Zinkevich and
  Balch}{2001}]{Zinkevich2001SymmetryLearning}
Martin Zinkevich and Tucker Balch.
\newblock Symmetry in markov decision processes and its implications for single
  agent and multi agent learning.
\newblock In {\em In Proceedings of the 18th International Conference on
  Machine Learning}. Citeseer, 2001.

\end{thebibliography}
%------------------------------------------------------------------------------------------------------------------
\end{document}